\pdfoutput=1

\documentclass[11pt]{article}

\usepackage[final]{acl}

\usepackage{times}
\usepackage{latexsym}

\usepackage[T1]{fontenc}

\usepackage[utf8]{inputenc}

\usepackage{microtype}

\usepackage{inconsolata}

\usepackage{graphicx}

\usepackage{url}
\usepackage{makecell}
\usepackage{epstopdf}
\usepackage{tipa}
\usepackage{hyperref}
\usepackage{xstring}
\usepackage{amsfonts}
\usepackage{amsmath}
\usepackage[T1]{fontenc}
\usepackage{booktabs}
\usepackage{arabtex}

\usepackage{hhline}
\usepackage{pdfpages}
\usepackage{utf8}
\newcommand{\hide}[1]{}

\newcommand{\AMADDA}{{\={A}}}
\newcommand{\AHAMZAUP}{{\^{A}}}
\newcommand{\WHAMZA}{{\^{w}}}
\newcommand{\AHAMZADN}{{\v{A}}}
\newcommand{\YHAMZA}{{\^{y}}}
\newcommand{\TAMARBUTA}{{$\hbar$}}
\newcommand{\TAMAR}{{$\hbar$}}

\newcommand{\SHIN}{{\v{s}}}
\newcommand{\ZA}{{\v{D}}} 
\newcommand{\AYN}{{$\varsigma$}}
\newcommand{\GAYN}{{$\gamma$}}

\newcommand{\SHADDA}{{$\sim$}}
\newcommand{\DAGGER}{{\'{a}}}
\newcommand{\AWASLA}{{\"{A}}}

\usepackage{multirow}
\usepackage{amssymb}
\usepackage[super]{nth}
\usepackage{arydshln}
\usepackage{lipsum}
\usepackage{bm}
\usepackage{caption}

\usepackage{gb4e}\noautomath
\usepackage{setspace}

\usepackage[colorinlistoftodos,textsize=tiny]{todonotes}

\newcommand{\camelprop}{\textsc{CamelProp}}
\newcommand{\camelwiki}{\textsc{CP-Wiki}}
\newcommand{\camelsama}{\textsc{CP-Sama}}
\newcommand{\camelwikiD}{\textsc{CP-Wiki-D3K}}

\usepackage{stackengine}
\usepackage{tikz}
\newcommand\dottedcircle{\tikz \draw [line cap=round, line width=0.15ex, dash pattern=on 0pt off 1.95\pgflinewidth] (0,0) circle [radius=0.6ex];}
\makeatletter
\newcommand{\fatha}{%
\bgroup \set@arabfont
\stackon[0.5pt]{\dottedcircle}{\char'013}%
\egroup%
}
\newcommand{\kasra}{%
\bgroup \set@arabfont
\stackunder[0.5pt]{\dottedcircle}{\char'013}%
\egroup%
}
\newcommand{\damma}{%
\bgroup \set@arabfont
\stackon[0.5pt]{\dottedcircle}{\char'014}%
\egroup%
}
\newcommand{\fathatan}{%
\bgroup \set@arabfont
\stackon[0.5pt]{\dottedcircle}{\char'023}%
\egroup%
}
\newcommand{\kasratan}{%
\bgroup \set@arabfont
\stackunder[0.5pt]{\dottedcircle}{\char'023}%
\egroup%
}
\newcommand{\shaddah}{%
\bgroup \set@arabfont
\stackon[0.5pt]{\dottedcircle}{\char'017}%
\egroup%
}
\newcommand{\dammatan}{%
\bgroup \set@arabfont
\stackon[0.5pt]{\dottedcircle}{\char'024}%
\egroup%
}
\newcommand{\skun}{%
\bgroup \set@arabfont
\stackon[0.5pt]{\dottedcircle}{\char'025}%
\egroup%
}
\newcommand{\alif}{%
\bgroup \set@arabfont
\stackon[0.5pt]{\dottedcircle}{\char'027}%
\egroup%
}

\setcode{utf8}
\vocalize

\title{Proper Noun Diacritization for Arabic Wikipedia:\\ A Benchmark Dataset}

\newcommand*{\authormark}[1][*]{\textsuperscript{#1}}
\author{
    Rawan Bondok,\authormark[1]
    Mayar Nassar,\authormark[1,2]
    Salam Khalifa,\authormark[1,3]
    Kurt Micallef,\authormark[1,4]
    Nizar Habash\authormark[1]\\
    {\normalfont Computational Approaches to Modeling Language (CAMeL) Lab} \\
    {\normalfont\authormark[1]New York University Abu Dhabi,}
    {\normalfont\authormark[2]Ain Shams University}\\
    {\normalfont\authormark[3]Stony Brook University,}
    {\normalfont\authormark[4]Department of Artificial Intelligence, University of Malta}\\
    {\normalfont\texttt{\{rawan.bondok,nizar.habash\}@nyu.edu}, \texttt{mayar.nassar@art.asu.edu.eg}}\\
    {\normalfont\texttt{salam.khalifa@stonybrook.edu}, \texttt{kurt.micallef@um.edu.mt}}
}

\usepackage[utf8]{inputenc} 
\usepackage[T1]{fontenc}
\begin{document}
\maketitle
\begin{abstract}
Proper nouns in Arabic Wikipedia are frequently undiacritized, creating ambiguity in pronunciation and interpretation, especially for transliterated named entities of foreign origin. While transliteration and diacritization have been well-studied separately in Arabic NLP, their intersection remains underexplored. In this paper, we introduce a new manually diacritized dataset of Arabic proper nouns of various origins with their English Wikipedia equivalent glosses, and present the challenges and guidelines we followed to create it. We benchmark GPT-4o on the task of recovering full diacritization given the undiacritized Arabic and English forms, and analyze its performance. Achieving 73\% accuracy, our results underscore both the difficulty of the task and the need for improved models and resources.  We release our dataset to facilitate further research on Arabic Wikipedia proper noun diacritization.\footnote{\url{https://github.com/CAMeL-Lab/CamelProp}\label{footCP}}
\end{list} 
\end{abstract}


\section{Introduction}

\begin{figure}[t]
    \centering
  \includegraphics[width=0.9\columnwidth]{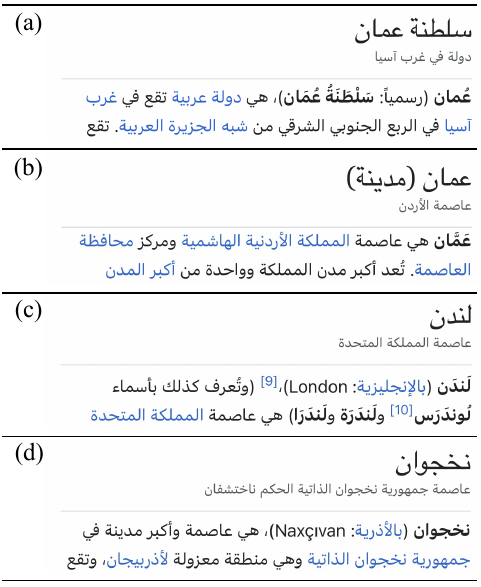}
    \caption{Four Arabic Wikipedia entries: (a) \<عمان> \textit{{\AYN}mAn} `Oman', (b) \<عمان> \textit{{\AYN}mAn} `Amman', (c) \<لندن> \textit{lndn} `London', and (d) \<نخجوان> \textit{nxjwAn} `Nakhchivan'. 
    All titles lack diacritics.
    Lead sentences do not consistently use diacritics: 
    (a) \textit{{\AYN}\textbf{u}mAn}, (b) \textit{{\AYN}\textbf{a}m\textbf{{\SHADDA}a}An}, and  (c) \textit{l\textbf{a}nd\textbf{a}n}; but (d) lacks diacritics, allowing multiple readings.}
    \label{fig:example}
\end{figure}

Arabic Wikipedia, like other language editions, has been a valuable resource for both its readers and NLP research. In this paper, we focus on a particular limitation rooted in Arabic's abjad orthography, where diacritics are typically omitted \cite{elgamal-etal-2024-arabic} except for children's books and religious texts.
This omission leads to ambiguity in pronunciation and interpretation, especially for proper nouns.  Some Arabic Wikipedia articles address this issue by providing partial or full diacritization in their lead sentences. For instance, \<عمان>~\textit{{\AYN}mAn}\footnote{Arabic HSB Romanization \cite{Habash:2007:arabic-transliteration}.} can refer to either  \<عُمَان>~\textit{{\AYN}umaAn} `Oman' or  \<عَمَّان> \textit{{\AYN}am{\SHADDA}aAn} `Amman' depending on the diacritization (Figure~\ref{fig:example}).  But more often than not, these diacritics are missing. In our dataset we found 99.45\% of all entries had no diacritics.  Our intention is to solve this limitation.

The work presented in this paper lies at the intersection of three commonly but often independently studied Arabic NLP tasks: \textit{transliteration}, \textit{diacritization}, and \textit{lemmatization}. 

\textbf{Transliteration} is the mapping of words, primarily proper nouns, from one script to another, usually in the context of machine translation \cite{Beesley:1997:romanization,benites-etal-2020-translit,chen-etal-2018-report}. It poses challenges due to misalignments between scripts, differences in representing phonology and morphology, and historical ad hoc conventions.

\textbf{Diacritization}, or diacritic restoration, aims at recovering omitted diacritics in languages that rely on them for disambiguation \cite{alqahtani-etal-2019-efficient,darwish-etal-2017-arabic,Abandah:2015:automatic}. While both transliteration and diacritization have been well studied for Arabic, they are typically treated in isolation. An exception is the work of \citet{MUBARAK09.81}, which considers both in the context of Arabic to English proper noun transliteration.

\textbf{Lemmatization} maps inflected words to their base forms. This is particularly important for morphologically rich languages such as Arabic \cite{Roth:2008:arabic}. In the context of Wikipedia entries, providing the lemmas is useful to readers as it gives them a grounding on how to interpret and later inflect the word forms properly.

More concretely, we focus here on mapping pairs of undiacritized Arabic proper nouns and their English glosses to fully diacritized and lemmatized Arabic forms. The task can be viewed as partial transliteration, where Roman-script vowels help infer (or transliterate into) Arabic diacritical marks. For example, \<نخجوان> \textit{nxjwAn} `Nakhchivan' (from Figure~\ref{fig:example}) should ideally be mapped to \<نَخْجِوَان> \textit{nax.jiwaAn}, rather than incorrect alternatives like \<نِخْجَوَان> \textit{nix.jawaAn} or  \<نُخْجِوَان> \textit{nux.jiwaAn}.

We present a new dataset of 3,000 unique Arabic Wikipedia proper nouns annotated with gold lemma-level diacritizations. Each entry is paired with its English Wikipedia equivalent, enabling the study of joint diacritization and transliteration. We benchmark GPT-4o \cite{openai2024gpt4technicalreport}, which shows promising results but struggles with spelling variants and ambiguity. The dataset covers a range of named entities (people, places, and organizations) and includes 3,362 total pairs to reflect multiple valid diacritizations based on the gloss.

Our contributions are:
\begin{itemize}
    \item A publicly available gold-standard dataset of Arabic Wikipedia proper nouns with English equivalents.\footref{footCP}
    \item A GPT-4o benchmark and detailed error analysis for Arabic proper noun diacritization.
\end{itemize}

The remainder of this paper is structured as follows. Section~\ref{sec:ling} outlines Arabic linguistic aspects. Section~\ref{sec:related} reviews related work. Sections~\ref{sec:data} and \ref{sec:annotation} describe our dataset and annotation process. Section~\ref{sec:eval} presents evaluation results and error analysis.

\section{Linguistic Background}
\label{sec:ling}

\begin{table}[t!]
\centering
\begin{tabular}{lccc} 
\toprule
\textbf{Diacritic} & \multicolumn{3}{c}{\textbf{Example}} \\
\midrule
Fatha               & \<بَ>     & \textit{ba}    & {/ba/} \\
Damma               & \<بُ>     & \textit{bu}    & {/bu/} \\
Kasra               & \<بِ>     & \textit{bi}    & {/bi/} \\

Shadda              & \<بّ>     & \textit{b{\SHADDA}}    & {/bb/} \\
Sukun               & \<بْ>     & \textit{b.}     & {/b/} \\
Dagger Alif         & \<بٰ>     & \textit{b{\DAGGER}}    & {/ba:/} \\
Shadda + Fatha      & \<بَّ>     & \textit{b{\SHADDA}a}   & {/bba/} \\
Shadda + Damma      & \<بُّ>     & \textit{b{\SHADDA}u}   & {/bbu/} \\
Shadda + Kasra      & \<بِّ>     & \textit{b{\SHADDA}i}   & {/bbi/} \\
\midrule
Long vowel /a/      & \<بَا>     & \textit{baA}    & {/ba:/} \\
Long vowel /u/      & \<بُو>     & \textit{buw}    & {/bu:/} \\
Long vowel /i/      & \<بِي>     & \textit{biy}    & {/bi:/} \\
Shadda + Long vowel /a/      & \<بَّا>     & \textit{bbaA}    & {/bba:/} \\
Shadda + Long vowel /u/      & \<بُّو>     & \textit{bbuw}    & {/bbu:/} \\
Shadda + Long vowel /i/      & \<بِّي>     & \textit{bbiy}    & {/bbi:/} \\
\midrule
Glide w     & \<بَوْ>     & \textit{baw.}    & {/baw/} \\
Glide y      & \<بَيْ>     & \textit{bay.}    & {/bay/} \\

\bottomrule
\end{tabular}
\caption{Examples of Arabic diacritics, their transliterations, and phonological values. We exclude nunation diacritics as they are not used in our lemmas.}
\label{tab:diacritics}
\end{table}


\begin{table*}[t!]
\tabcolsep8pt
\centering
\begin{tabular}{rllrll}
\toprule
\multicolumn{2}{c}{\textbf{Input Arabic}} & \textbf{Gloss} & \multicolumn{2}{c}{\textbf{Lemma Arabic}} & \textbf{Transformation} \\
\midrule
\<الست> & \textit{Alst} & Al-Sit & \<سِتّ> & \textit{sit{\SHADDA}} & DET → $\phi$ \\
\<الواس> & \textit{AlwAs} & Elvas & \<إِلْوَاس> & \textit{{\AHAMZADN}il.waAws} & Bare Alif → Alif Hamza \\
\<العجم> & \textit{Al{\AYN}jm} & Al-Ajam & \<عَجَم> & \textit{{\AYN}ajam} & DET → $\phi$ \\
\<الغظاة> & \textit{Al{\GAYN}{\ZA}A{\TAMARBUTA}} & Al-Ghadhah &  \<غَظَاة> & \textit{{\GAYN}a{\ZA}aA{\TAMARBUTA}} & DET → $\phi$ \\
\<فنزويليون> & \textit{fnzwylywn} & Venezuelans & \<فِنِزْوِيلِيّ> & \textit{finiz.wiyliy{\SHADDA}} & 3MP → $\phi$ \\
\<الجيبوتيون> & \textit{Aljybwtywn} & Djiboutians & \<جِيبُوتِيّ> & \textit{jiybuwtiy{\SHADDA}} & DET+3MP → $\phi$ \\
\bottomrule
\end{tabular}
\caption{Examples of lemmatization transformations from Arabic input (inflected) words to canonical lemmas, with English glosses and corresponding changes.} 
\label{tab:lemmatizations}
\end{table*}

\subsection{Arabic Diacritization}\label{sec:ling_diac}

Arabic orthography follows an \textit{Abjad} system \cite{daniels2013arabic}, where letters encode consonants and diacritical marks represent short vowels, nunation (case endings), gemination, and vowel absence. Diacritic clusters are typically limited to a Shadda ({\shaddah}~{\SHADDA}) followed by a short vowel or nunation diacritic.
Three letters, \<ا>~\textit{A}, \<و>~\textit{w}, and \<ي>~\textit{y} (henceforth \textit{AWY}), encode long vowels when preceded by a matching short vowel and not followed by any diacritic: \<ا>{\fatha} \textit{aA} (/a:/), \<و>{\damma} \textit{uw} (/u:/), and \<ي>{\kasra} \textit{iy} (/i:/). These letters are often used with foreign name transliterations to mark the vowel quality independent of length, e.g., \<بين> \textit{byn} `Ben' or `Bean'.
The letters \<و> and \<ي> also serve as glides (/w/ and /y/) when preceded by {\fatha}~\textit{a} and followed by a sukun ({\skun}~\textit{.}). The letter \<ا>~\textit{A} functions as a carrier for initial short vowels (\textit{Alif Wasla}, \<ٱ>~\textit{\AWASLA}).
Additionally, Arabic uses letters with attached Hamza diacritics, e.g., \<أ>~\textit{\AHAMZAUP}, \<إ>~\textit{\AHAMZADN}, \<آ>~\textit{\AMADDA}, \<ؤ>~\textit{\WHAMZA}, and \<ئ>~\textit{\YHAMZA}. The omission of Hamzas is treated as a spelling error and corrected during diacritization. 

See Table~\ref{tab:diacritics} for examples, and \newcite{darwish-etal-2017-arabic} and \newcite{elgamal-etal-2024-arabic} for more details on Arabic diacritics.

\subsection{Arabic Lemmatization}\label{sec:ling_lemma}
In Arabic morphology, the lemma is the canonical form (also known as citation form) of a word that abstracts over its inflected variants, including gender, number, person, and case, as well as attached clitics \cite{Roth:2008:arabic,Habash:2010:introduction}. 
Table~\ref{tab:lemmatizations} shows examples of input forms and their corresponding lemmas.
In our context, lemmatization is simpler than in free-form text: we focus only on proper nouns, an English gloss is available to guide vowelization, and clitics are rare. 
The main challenges are distinguishing between baseword and determiner uses of \<ال>~\textit{Al}~(DET) initial substring (see Table~\ref{tab:lemmatizations}~rows 1-2), and handling plural endings~(3MP) \<ون>~\textit{uw} in demonyms (Table~\ref{tab:lemmatizations}  rows 5-6).

\begin{table}[t!]
\tabcolsep4pt
\centering
\begin{tabular}{cccc}
\toprule
 & \textbf{Pronunciation} & \textbf{Arabic} & \textbf{Transliteration} \\
\midrule
(a) & /bla:stik/     & \<بْلَاسْتِك>     & \textit{b.laAs.tik} \\
(b) & /bila:stik/    & \<بِلَاسْتِك>     & \textit{bilaAstik} \\
(c) & /bla:stik/     & \<بْلَاسْتِيك>    & \textit{b.laAs.tiyk} \\
(d) & /bila:stik/    & \<بِلَاسْتِيك>    & \textit{bilaAstiyk} \\
(e) & /bla:sti:k/    & \<بْلَاسْتِيك>    & \textit{b.laAs.tiyk} \\
(f) & /bila:sti:k/   & \<بِلَاسْتِيك>    & \textit{bilaAstiyk} \\
(g) & /bala:sti:k/   & \<بَلَاسْتِيك>    & \textit{balaAstiyk} \\
(h) & /ibla:stik/    & \<اِبْلَاسْتِيك>   & \textit{Aib.laAs.tiyk} \\
\bottomrule
\end{tabular}
\caption{Variants of the pronunciation and transliteration of the Arabic word for `plastic'. Three basic spellings: (a-b)~\<بلاستك>~\textit{blAstk}, (c-g)~\<بلاستيك>~\textit{blAstyk}, and (h)~\<ابلاستك>~\textit{AblAstk}, with various diacritizations.} 
\label{tab:plastic_variants}
\end{table}

\subsection{Arabic Transliterations}
Transliteration from Roman script to Arabic script presents several challenges, primarily due to the misalignment between the phonology of the original language and its Roman script orthography, as well as differences between the phonology of the original languages and Arabic. Arabic, for example, has fewer vowels (6 in Arabic vs. 15 in English), and some missing (no /p/ or /v/) and additional consonants (e.g., emphatic /d/ and /q/).
Arabic dialects vary in phonology, including sound quality, letter mapping, and syllabification, leading to multiple valid transliterations. For instance, the borrowed word `plastic' can have different pronunciations and spellings, reflecting variations in vowels and syllabification (see Table~\ref{tab:plastic_variants}).
%
%
During annotation, we followed Wikipedia spelling and aligned with the English gloss. The team included Egyptian, Sudanese, and Levantine speakers, with an Egyptian speaker as the primary annotator.


\section{Related Work}
\label{sec:related}
\subsection{Diacritization in Arabic NLP}

Arabic diacritization has been extensively studied using both statistical and neural methods. Some approaches treat it as a standalone task \citep{Zitouni:2006:maximum, Mubarak:2019:highly}, while others integrate it into multitask learning frameworks alongside linguistically related tasks such as part-of-speech tagging \citep{habash-rambow-2005-arabic, alqahtani-etal-2020-multitask}.

A commonly adopted strategy involves the use of morphological analyzers. For instance, Camelira (Camel~Tools) implements an analyze-and-disambiguate pipeline: a morphological analyzer generates candidate analyses, which are then ranked by a classifier \cite{obeid-etal-2020-camel,obeid-etal-2022-camelira}. Similarly, Farasa uses morphological patterns to diacritize words \cite{darwish-etal-2017-arabic}. 

Systems such as Farasa and Camel~Tools have demonstrated strong performance on sentence-level diacritization tasks.
However, these systems are not directly applicable to our task, which centers on isolated proper nouns, adheres to a task-specific diacritization schema, and incorporates lemma mapping. Unlike sentence-based systems that leverage surrounding context for disambiguation, our task involves context-free diacritization, which poses distinct challenges (see Section~\ref{subsec:guidelines}).

\subsection{Lemmatization in Arabic NLP}

Lemmatization is another core task in Arabic NLP, and several tools offer robust performance across a variety of syntactic categories \citep{obeid-etal-2020-camel,obeid-etal-2022-camelira,JARRAR2024378}. However, our lemmatization task has a narrower scope: it is limited to proper nouns that have a limited inflectional space (see Section~\ref{subsec:guidelines} for further details on our lemmatization space).

\subsection{Transliteration in Arabic NLP}

Earlier research on Arabic–English transliteration relied on statistical approaches \citep{article}, followed by more targeted work on proper nouns using models such as phonemic memory networks \cite{tian-etal-2022-improving-english}. A persistent challenge in this area is the lack of standardization in transliterating foreign names into Arabic, a problem exacerbated by the omission of diacritics \cite{001931ar,1bd0c6cd-2a8a-36f4-8a46-0b6886b6c132}.

To address the lack of standardization and limited resources, we introduce a new dataset and annotation guidelines specifically designed for the task of utilizing proper noun transliteration as a signal for Arabic diacritization.

Prior efforts investigated the intersection of transliteration and diacritization, such as \citet{MUBARAK09.81} and \citet{darwish-etal-2017-arabic}. \citet{MUBARAK09.81} used diacritization as a preprocessing step to transliteration. Although, the approach presented in \citet{darwish-etal-2017-arabic} for automatically diacritizing transliterated words included leveraging English transliterations to generate Arabic diacritized proper nouns, both their training and test sets were limited in size (500 and 200 instances, respectively). Our resource, in contrast, is publicly available, much larger (3,000 diacritized lemmas), and benchmarked for robust evaluation and development.

\subsection{Arabic Proper Noun Resources}

Although various Arabic proper noun datasets exist, they often suffer from limited accessibility, lack of diacritics, or domain constraints. For example, \citet{matthews2007transliteration} compiled a list of 10,001 Arabic names, but the dataset is not publicly available. \citet{eryani-habash-2021-automatic} provide automatically Romanized Arabic bibliographic entries without diacritics, and both the Dan database \citep{halpern2009lexicon} and SAMA \citet{Graff:2009:standard} include diacritized proper nouns, but they were mainly collected form news sources.

\newcite{khairallah-etal-2024-camel} released a large set of proper nouns as part of their CamelMorph Arabic morphological analyzer (henceforth {\camelprop}, CP for short). 
The dataset consists of two distinct portions: (a) {\camelsama}, which extends the SAMA \cite{Graff:2009:standard} proper-noun list and updates their diacritizations; and (b) {\camelwiki} which comprises 63K entries extracted from a Wikidata dump (14-Mar-2023).\footnote{\url{https://dumps.wikimedia.org/wikidatawiki/entities/}}  The {\camelwiki} was filtered by \newcite{khairallah-etal-2024-camel} to include only single word entities in Arabic and English, and covering only personal and family names, locations and organizations.  Unfortunately, \newcite{khairallah-etal-2024-camel} did not provide diacritizations for the {\camelwiki} portion. Our interest in this topic started by this problem in their open-source resource, which was not usable for our purposes. We discuss these datasets further in Section~\ref{sec:data}.

In this work, we present the first publicly available dataset of maximally manually diacritized 
and lemmatized Arabic proper nouns on a portion of the {\camelwiki} dataset sourced from Wikimedia and manually annotated using English equivalents in a consistent and standardized annotation scheme. To support future work, we also release detailed annotation guidelines and provide the first benchmark of GPT-4o’s performance on this task, offering a new resource for evaluating Arabic proper noun diacritization and transliteration.



\begin{table*}[t!h!]
\tabcolsep3pt
\centering
\begin{tabular}{lccc}
\toprule
 & \textbf{{\camelsama}} & \textbf{{\camelwiki}} & \textbf{{\camelwikiD}} \\
\midrule
\textbf{Unique Arabic} & 6,022 & 63,417 & 3,000 \\
\textbf{Arabic-English Entries} & 7,202 & 71,251 & 3,362 \\
\textbf{English glosses per entry} & 1.20 & 1.12 & 1.12 \\
\textbf{Average Freq} & 205,077 & 97,438 &61,544 \\
\textbf{Median Freq} & 11,732  & 87 & 75 \\
\textbf{Average Freeman Score} &  0.92  &0.91 &  0.91 \\
\textbf{Diacritizations} & Yes & No & Yes\\

\bottomrule
\end{tabular}
\caption{Comparison of dataset statistics across {\camelsama}, {\camelwiki}, and the annotated subset {\camelwikiD}.}
\label{tab:araprop_comparison}
\end{table*}


\begin{table*}[t!]
       \centering
       \tabcolsep10pt
    \begin{tabular}{lcc}
     \toprule
        \textbf{Class} & \textbf{{\textsc{CP-Wiki}}}
         & \textbf{{\textsc{CP-Wiki-D3K}}}  \\
         \midrule
         \textbf{Location} & 77.1\% & 85.2\% \\
         \textbf{Name} & 25.5\% & 35.0\% \\
         \textbf{Organization} & 2.0\% & 2.0\% \\
          \bottomrule
    \end{tabular}
    \caption{Distribution of different named entity classes across {\camelwiki} and {\camelwikiD}}
\label{tab:class_distribution}
\end{table*}

\section{Datasets}
\label{sec:data}

We work with the {\camelprop} dataset, released as part of CamelMorph, an Arabic morphological analyzer, by \newcite{khairallah-etal-2024-camel}. As noted in Section~\ref{sec:related}, it consists of two parts: {\camelsama} and {\camelwiki}. 
We randomly selected 3,000 unique Arabic-script proper nouns from {\camelwiki} for manual annotation, forming our dataset {\camelwikiD}.

Table~\ref{tab:araprop_comparison} compares the three datasets in terms of unique Arabic entries and full Arabic–English gloss pairs, average and median frequency and Arabic-English phonological similarity. For frequency we used the Arabic Frequency list from \citet{Khalifa:2021:Camel_Frequency}. For phonological similarity, we used the Freeman similarity score \cite{Freeman:2006:cross}. The original data included multiple glosses per Arabic word (12–20\% extra on average). We normalized this by splitting them into separate one-to-one pairs. For example, \<آنا> \textit{{\AMADDA}nA}, glossed as `A'ana; Ana; Anna', became three distinct entries:  (\<آنا> \textit{{\AMADDA}nA}, `A'ana'), (\<آنا> \textit{{\AMADDA}nA}, `Ana'), and (\<آنا> \textit{{\AMADDA}nA}, `Anna'). 
Thus, our 3,000 Arabic words expanded to 3,362 Arabic–gloss pairs.
While phonological similarity is only slightly lower in {\camelwikiD} and {\camelwiki}, the overall frequency in {\camelwiki} and {\camelwikiD} is significantly lower than {\camelsama}, highlighting the importance of modeling the diacritization of low-frequency proper nouns in Wikipedia and NLP.

In addition to frequency and phonological similarity, we examined the distribution of named entity categories, namely, personal and family names, locations, and organizations, across both the original {\camelwiki} dataset and the manually annotated subset, {\camelwikiD}. The distributions were broadly similar, with location entities being the majority in both ({\camelwiki}: 77.1\%, {\camelwikiD}: 85.2\%), followed by names and organizations. This consistency supports the representativeness of {\camelwikiD} for studying diacritization across entity types. Table \ref{tab:class_distribution} reports the detailed percentage breakdown of entity classes in both datasets.

\begin{table*}[ht]
\centering
\tabcolsep5pt
\begin{tabular}{rlllrl}
\toprule
\multicolumn{2}{c}{\textbf{Invalid Lemma}} & \textbf{Gloss} & \textbf{Issue} & \multicolumn{2}{c}{\textbf{Corrected Lemma}} \\
\midrule
\<سانشِيز> & \textit{sAn{\SHIN}iyz} & Sanchez & Long vowels require preceding 
diacritics & \<سَانْشِيز> & \textit{saAn.{\SHIN}iyz} \\
\<كَرَمُ> & \textit{karamu} & Karam & Final letter cannot have a diacritic & \<كَرَم> & \textit{karam} \\
\<عَضُّوم> & \textit{{\AYN}aDu{\SHADDA}wm} & Addoum & Short diacritic cannot precede Shadda & \<عَضُّوم> & \textit{{\AYN}aD{\SHADDA}uwm} \\

\bottomrule
\end{tabular}

\caption{Examples of malformed words and their corrected lemmas with transliterations.}
\label{tab:well-formed}
\end{table*}

\section{Data Annotation}
\label{sec:annotation}

In this section, we discuss the diacritization guidelines we used, as well as the setup for initial automatic processing followed by manual correction.

\subsection{Diacritization Guidelines}
\label{subsec:guidelines}

We follow the Arabic maximal diacritization guidelines as presented in \newcite{elgamal-etal-2024-arabic} with a small number of modifications to fit the purpose of our task. We list the most important decisions that are different from standard Arabic diacritization.

\paragraph{The Lemmatization Requirement}
This effort focuses exclusively on the diacritization of proper nouns and mapping them to their lemmas. As such, we require the removal of clitics such as the definite article and the removal of plural suffixes (see Section~\ref{sec:ling_lemma}).

\paragraph{Input Spelling Integrity} Aside from the minimal changes connected to lemmatization, and corrections of the obligatory Hamza diacritic in Alif Hamza forms (see Section~\ref{sec:ling_diac}), we do not add, remove, or modify any letters in the provided input.

\paragraph{Consonant Clusters in Foreign Names}  
While standard Arabic generally avoids consonant clusters, our dataset includes many foreign proper nouns where such clusters are phonetically natural. To more faithfully capture their pronunciation, we allowed forms with consecutive consonants, either multiple letters marked with Sukuns, or a Sukun followed by a letter with Shadda (geminated), even though this departs from Standard Arabic diacritization norms.
For example,  \<إلكتريك> \textit{{\AHAMZADN}lktryk} `Electric' should be diacritized as \<إِلِكْتْرِيك> \textit{{\AHAMZADN}ilik.t.riyk} (with the consonant cluster /tr/), and 
\<زدينيك>~\textit{zdynyk} `Zdeněk' should be diacrtized as \<زْدِنِيك>   \textit{z.diniyk} with initial /zd/ cluster.

\paragraph{Final Letter Ya}
The final letter \<ي>~\textit{y} has multiple diacritizations 
that overlap with changes in dialectal Arabic, i.e. the softening of final y-gemination into /i/.
As such, we had to dedicate part of the guidelines to outline the rules for diacritizing it as a geminated /yy/, a long vowel /i:/ or a glide /ay/.

The geminated version is the most specific in requirements with three possible cases:
\begin{itemize}
    \item The gemination comes from the root or pattern of the word such as the final Ya in \<رَخِيّ> \textit{raxiy{\SHADDA}} `Ar-Rakhi'.
    \item The lemma can be interpreted as having the derivational attribution suffix Ya-Nisba, e.g., \<إِشْبِيلِيّ> \textit{{\AHAMZADN}i{\SHIN}.biyliy{\SHADDA}} `Sevillian' (of or related to \<إِشْبِيلِيَّة> \textit{{\AHAMZADN}i{\SHIN}.biyliy{\SHADDA}a{\TAMARBUTA}} `Seville').
     \item  
     Gemination is necessary to reflect the pronunciation of certain foreign names, such as \<أَرْكْوِيّ>~\textit{{\AHAMZAUP}ar.k.wiy{\SHADDA}} `Arcueil'.

    \end{itemize}

For other cases, if the final vowel sounds like a short /i/ or a long /i:/ and has a corresponding \<ي>~\textit{y}, it is diacritized resembling a long vowel, e.g., \<أغاسي> \textit{{\AHAMZAUP}gAsy} Agassi, should be diacritized as \<أَغَاسِي> \textit{{\AHAMZAUP}agaAsiy}.
The glide version is straightforward as it has a distinct phonological signal. One example is the word   
\<نَي> \textit{nay} `Ney'.

\paragraph{Checking Well-formedness}
To ensure consistency with our annotation guidelines, we implemented automated checks to validate the well-formedness of diacritized lemmas. While these checks do not guarantee correctness, they are effective at identifying common errors and inconsistencies.  We use these checks on both human and automatic annotations.
See Table~\ref{tab:well-formed} for examples.
 
\subsection{Initial Automatic Diacritization}
\label{sec:postprocessing}  
To speed up the annotation process, we gave our annotator an automatically diacirtized version of the data. We used  GPT-4o with Arabic Input and English Gloss (comparable to the best setting in Section~\ref{sec:eval}). 
 At the time of generating the initial automatic diacritization, we considered this a reasonable starting point.

\begin{table*}[t]
    \centering
    \tabcolsep4pt
    \begin{tabular}{lrlrlrl}
    \toprule
    \textbf{Type of Disagreement} & \textbf{Freq}  & \textbf{Gloss}& \multicolumn{2}{c}{\textbf{First Annotator}} & \multicolumn{2}{c}{\textbf{Second Annotator}} \\
    \midrule
    
    Kasra $\leftrightarrow$ Sukun & 13 & Tibet & \<تِبِت> & tibit & \<تِبْت> & tib.t  \\
    Kasra $\leftrightarrow$ Fatha & 9 & Shechem & \<شِكِيم> & {\SHIN}ikiym & \<شَكِيم> &{\SHIN}akiym  \\
    Consonant $\leftrightarrow$ Long vowel & 9 & Jane & \<جَيْن> &jay.n & \<جِين> & jiyn\\
    Sukun $\leftrightarrow$ Damma & 5 & Acquaviva &\<أَكُوَافِيفَا> & {\AHAMZAUP}akuwaAfiyfaA & \<أَكْوَافِيفَا> &{\AHAMZAUP}ak.waAfiyfaA \\
    Sukun $\leftrightarrow$ Fatha & 1 & Aminadav & \<عَمِينْدَاف> & {\AYN}amiyn.daAf & \<عَمِينَدَاف> &{\AYN}amiynadaAf \\
    Shadda $\leftrightarrow$ $\phi$ & 1 & Oss&\<أُوسّ> & {\AHAMZAUP}uws{\SHADDA}& \<أُوس> & {\AHAMZAUP}uws \\
    \bottomrule
    \end{tabular}
    \caption{Types of disagreements in Inter-Annotator Evaluation}
    \label{tab:IAA}
\end{table*}

\paragraph{GPT-4o postprocessing}
The output of GPT-4o was not always usable as is. 
When applying well-formedness checks to the diacritized outputs generated by GPT-4o, we observed several recurring patterns of errors that compromised the validity of the diacritized forms. In response, we developed an automated pipeline specifically aimed at correcting these systematic errors.\footref{footCP}
The automatic correction procedures included the following operations:

\begin{itemize} 
\item Insertion of Fatha before Alif (\<ا> \textit{A}). 
 \item  Insertion of Kasra after Alif-Hamza-Below (\<إ>~\textit{\AHAMZADN}).   
\item  Normalization of Shadda-Vowel clusters such that the vowel diacritic follows the Shadda diacritic.
\item  Removal of final diacritics as lemmas do not have them.
\item Insertion of missing Sukuns to indicate vowel absence at the end of syllable or in a consoant cluster.
\item Removal of Fatha after Alif Madda (\<آ>  \textit{\AMADDA}).
\item  Mapping Non-Arabic Arabic-script letters, such as those used in Urdu or Persian, to their closest Arabic language form.   
\end{itemize}

\subsection{Manual Diacritization}
The manual diacritization and quality checks were carried out by a native speaker of Arabic from Egypt who is a trained linguist and a highly experienced annotator.
The annotation process initially was done in tandem with the finalization of the guidelines with a team of the authors working jointly to optimize the quality of the annotation. 
The annotator was provided an Arabic word, along with its English gloss, and a proposed diacritization from GPT-4o after being refined by the automatic post-process described above.  The annotations were carried on Google Sheets in a very simple setup. The annotator reviewed the proposed diacritization making changes where needed in accordance to the guidelines. The annotator made changes to 909 proposed lemmas out of 3,362 ($\sim$27\%).
In 213 instances (6.3\% of all entries), there was a change connected with lemmatization: 74\% relative involved the Al determiner, 22.5\% a change in Alif-Hamza spelling, and 3.3\% involving the demonym plural ending.

\subsection{Inter-annotator Agreement}
To assess the quality of our annotation and the consistency of our guidelines, we conducted an inter-annotator agreement study. 
A second annotator, a native Arabic speaker from Egypt, independently re-annotated a subset of 500 randomly selected samples from the dataset, utilizing the same annotation process and adhering to the same guidelines as the first annotator. 
Out of the 500 samples, the annotators fully agreed on 462 instances and disagreed on 38, resulting in an inter-annotator agreement rate of 92.4\%. Table~\ref{tab:IAA} presents the various types of inter-annotator disagreements along with their corresponding frequencies. Each row in the table represents a type of disagreement where the annotators selected different diacritics for the same word. For example, the first row shows instances where either one of the annotators chose a Kasra while the other selected a Sukun.

\section{Evaluation}
\label{sec:eval}

\subsection{Experimental Setup}
We perform computational experiments to perform the task of diacritization of proper nouns.
For this, we prompt GPT-4o on all of the annotated dataset described in Section~\ref{sec:annotation}.
We prompt the model with different input formats to assess its capabilities while giving it different levels of information: the inputs and the number of examples shown to the model (shots). We used default settings for optional parameters (e.g., temperature, top\_p) from the gpt-4o-2024-11-20 snapshot.\footnote{\url{https://platform.openai.com/docs/api-reference/chat/create}}

\paragraph{Inputs}
The model is given a detailed description of the task to be performed.
Our main experiments reflect all the information given to our annotator, where we provide the model with both the Arabic Input and the English Gloss (\textbf{Arabic  + Gloss}).
Additionally, we also experiment with a more constrained setup where the model is provided solely with the the Arabic Input (\textbf{Arabic Only}).

\paragraph{Shots}
In addition to the different inputs, we also consider further experiments where we supply the model with varying number of examples to learn from\footref{footCP}
Hence, in addition to just providing the input to diacritize (\textbf{Zero-Shot}), we also supply the model with a single example (\textbf{One-Shot}), and 80 examples (\textbf{Few-Shot}).
The examples are randomly sampled from the {\camelsama} data.
The one-shot and few-shot examples were selected once and reused across all model prompts.
However, since {\camelsama} has fully lemmatized Arabic Inputs, we manually manipulated some of the examples to have a representation of clitic removal and Hamza normalization.
Refer to Appendix~\ref{app:prompts} for a more detailed description of the prompts used.

\paragraph{Post-processing}
As a post-processing step, the outputs were ran through the same processing pipeline mentioned in Section~\ref{sec:postprocessing}.
To evaluate the performance of the different experiments, we computed two metrics: accuracy by measuring the exact match between the post-processed output and the gold-standard diacritization and Levenshtein  edit distance \citep{Levenshtein:1966:binary} between the output and gold-standard diacritization.

\subsection{Results}\label{sec:results}

\begin{table}[t]
    \centering
    \begin{tabular}{lccc}
        \toprule
   \textbf{Input Format} &\textbf{Shots} &\textbf{Accuracy}& \textbf{Distance}  \\

        \midrule
        Arabic + Gloss &Zero  &         46.5\% 
        & 1.02\\
        Arabic + Gloss &One   &         61.9\%   
        & 0.64\\
         Arabic + Gloss &Few   & \textbf{73.0\%}
        & 0.41\\
        \midrule
        Arabic Only &Zero  &          36.7\% 
        & 1.29\\
        Arabic Only &One   &          49.7\%  
        & 0.86\\
         Arabic Only &Few   &         55.9\% 
         & 0.71\\
         \bottomrule  
    \end{tabular}
    \caption{GPT-4o model results on {\camelwikiD} in terms of exact match accuracy and Levenshtein edit distance.} 
    \label{tab:results_chatgpt}
\end{table}

\begin{table*}[t]
\tabcolsep3pt
\centering
\small
\begin{tabular}{rllrlrll}
\toprule
\multicolumn{2}{c}{\textbf{Input}} & \textbf{Gloss} & \multicolumn{2}{c}{\textbf{Reference}} & \multicolumn{2}{c}{\textbf{Prediction}} & \textbf{Error Type} \\
\midrule
\<العمود> & Al{\AYN}mwd & Al-Amud & \<عَمُود> & {\AYN}amuwd & \<عَمُود> & {\AYN}amuwd & Exact Match \\
\<أفراموفو> & ÂfrAmwfw & Avramovo & \<أَفْرَامُوفُو> & Âaf.raAmuwfuw & \<أَفْرَامُوفُو> & Âaf.raAmuwfuw & Exact Match \\
\<هاغن> & hA{\GAYN}n & Hagen & \<هَاغِن> & haA{\GAYN}in & \<هَاغِن> & haA{\GAYN}in & Exact Match \\
\midrule
\<إشتهارد> & ĂšthArd & Eshtehard & \<إِشْتِهَارْد> & Ăiš.tihaAr.d & \<إِشْتَهَارْد> & Ăiš.tahaAr.d & Diac \\
\<بلاجيفيتش> & blAjyfytš & Blažević & \<بْلَاجِيفِيتْش> & b.laAjiyfiyt.š & \<بِلَاجِيفِيتْش> & bilaAjiyfiyt.š & Diac \\
\<دسوق> & dswq & Desouk & \<دِسُوق> & disuwq & \<دُسُوق> & dusuwq & Diac \\
\midrule

\<ريبلاي> & ryblAy & Ripley & \<رِيبْلَاي> & riyb.laAy & \<رِيبْلِي> & riyb.liy & \textit{AWY} \\
\<ريكسينغين> & ryksyn{\GAYN}yn & Rexingen & \<رِيكْسِينْغِين> & riyk.siyn.{\GAYN}iyn & \<رِيكْسِنْغِن> & riyk.sin.{\GAYN}in & \textit{AWY} \\
\<جوندريزيك> & jwndryzyk & Gondrezick & \<جُونْدْرِيزِيك> & juwn.d.riyziyk & \<جُنْدْرِيزِيك> & jun.d.riyziyk & \textit{AWY} \\
\midrule
\<ميشيغان> & myšy{\GAYN}An & Michigan & \<مِيشِيغَان> & miyšiy{\GAYN}aAn & \<مِيشِيجَان> & miyšiyjaAn & \textit{j} $\leftrightarrow$ \textit{{\GAYN}} \\
\<تسيخانوف> & tsyxAnwf & Ciechanów & \<تِسِيخَانُوف> & tisiyxaAnuwf & \<تِسِيهَانُوف> & tisiyhaAnuwf & \textit{h} $\leftrightarrow$ \textit{x} \\
\<أردينة> & Ârdyn{\TAMAR} & Ardineh & \<أَرْدِينَة> & Âar.diyna{\TAMAR} & \<أَرْدِينَه> & Âar.diynah & {\TAMAR} $\leftrightarrow$ \textit{h}\\
\midrule
\<إيغيل> & Ăy{\GAYN}yl & Eagle & \<إِيغِيل> & Ăiy{\GAYN}iyl & \<إِيجِل> & Ăiyjil & Multiple \\
\<كرامة> & krAm{\TAMAR} & Gourrama & \<كُرَامَة> & kuraAma{\TAMAR} & \<كُورَامَا> & kuwraAmaA & Multiple\\
\<ايكوميديا> & AykwmydyA & Eco-Médias & \<إِيكُومِيدْيَا> & Ăiykuwmiyd.yaA & \<إِيكُومِيدْيَآس> & Ăiykuwmiyd.yaĀs & Multiple \\
\<بارافرانكا> & bArAfrAnkA & Barrafranca & \<بَارَّافْرَانْكَا> & baAr~aAf.raAn.kaA & \<بَارَافْرَانْكَة> & baAraAf.raAn.ka{\TAMAR} & Multiple \\
\bottomrule
\end{tabular}
\caption{Examples of evaluated instances along with their, reference and predicted diacritized forms, and corresponding error types. The error categories are diacritic mismatches (Diac), \textit{AWY} spelling changes (\textit{AWY}), several consonant and ta-marbuta substitutions (\textit{j} $\leftrightarrow$ \textit{{\GAYN}}, \textit{h} $\leftrightarrow$ \textit{x}, and  {\TAMAR} $\leftrightarrow$ \textit{h}), and those with multiple changes (Multiple).}
\label{tab:pred-errors}
\end{table*}

The results demonstrate that while diacritizing proper nouns remains a challenging task, incorporating the English gloss offers a valuable signal for the model. Notably, the best performance is achieved with few-shot, showing the effectiveness of providing a diverse and representative sample. Table~\ref{tab:results_chatgpt} shows the results with different prompts.

\subsection{Interplay of Frequency, Similarity, and Accuracy}
\label{sec:similarity-frequency-analysis}

We investigated how lexical frequency and phonological similarity \cite{Freeman:2006:cross} affect model performance under our best configuration: few-shot prompting with Arabic + Gloss.

The Freeman similarity score averaged a high 91\% across the dataset, consistent with the transliteration focus of the task. We binned the data into 10 intervals based on Freeman score. The lowest-similarity bins (up to 50\%), comprising only 3\% of the data, contained mostly high-frequency named entities and translations, e.g., \<مصر> \textit{mSr} for `Egypt' and \<عملاق > \textit{{\AYN}mlAq} for `jötnar'. Despite their low similarity, this group achieved 13.9\% higher accuracy and had, on average, 10 times the frequency compared to the rest of the data.
The bins up to 90\% similarity comprised 35\% of the data; their average frequency is only 5\% higher than the last bin, but their average accuracy is lower by 3.6\% absolute. 

We found strong negative correlation between accuracy and edit distance (-0.95), confirming that higher accuracy aligns with fewer character edits. Frequency and Freeman score showed a moderate negative correlation (-0.69), likely due to high-frequency translated names. Freeman similarity and accuracy were also moderately negatively correlated (-0.70), indicating that frequent but phonetically dissimilar words are still predicted accurately.

We analyzed performance across frequency quartiles (Q1 to Q4). Accuracy rose steadily from 65\% in Q1 to 80\% in Q4. The correlation between average frequency and accuracy across quartiles was 0.68, confirming the positive impact of frequency on model performance. Full analysis tables are presented in Appendix~\ref{app:freeman_frequency_analysis}.

\subsection{Error Analysis}
We analyzed errors from a randomly selected sample of 1,010 output entries from the best performing setup from Section~\ref{sec:results}, and classified errors into several categories based on observed patterns. There were 740 (73.3\%) exact matches (correct generations).

Of the 270 (26.7\%) errors, there were 175 cases where the error was only diacritization differences. See examples in Table~\ref{tab:pred-errors}. Upon further analysis of this class of errors, we found that the model overpredicts Fathas (+25\%) and Shaddas (+96\%), while underpredicting Kasras (-18\%) and Sukuns (-23\%), indicating imbalanced vowel modeling and overuse of gemination.	

The next largest class of errors, 60 cases, were those with spelling changes limited to the set of long vowel (and glides) letters \<ا> \textit{A}, \<و> \textit{w}, and \<ي> \textit{y} (\textit{AWY}). As we see in the examples in Table~\ref{tab:pred-errors}, the model has the tendency of dropping such letters rather than adding them. Another class of errors, 10 cases, are those with specific letter replacements such as \<ج> \textit{j} $\leftrightarrow$ \<غ> {\GAYN}, \<خ> \textit{x} $\leftrightarrow$ \<ه> \textit{h}, and \<ة> \textit{{\TAMAR}} $\leftrightarrow$ \<ه> \textit{h}. The final class of errors, 25 cases, are those with multiple changes happening at once. 

While these cases don't match the gold reference, they are plausible and acceptable alternatives in most cases, especially in the context of linguistic variation discussed in Section~\ref{sec:ling}. For example, the generated diacritization for \<بلاجيفيتش> \textit{blAjyfytš} `Blažević' as seen in Table~\ref{tab:pred-errors} (row 5), follows the common phenomena of breaking word initial complex onsets in many spoken dialects of Arabic and in MSA. Another example is the entry \<ايكوميديا> \textit{AykwmydyA} `Eco-Médias', where the input follows a pronunciation-based transliteration while the generated form adhered to the orthography of the gloss.


These variations highlight the need for modeling techniques and evaluation metrics that account for this  aspect of Arabic proper noun diacritization, which in turn requires additional annotated data.


\section{Conclusion and Future Work}

We presented a new 3,362 entry dataset of Arabic Wikipedia proper nouns annotated with gold-standard lemma diacritizations, paired with their English equivalents. This resource enables the joint study of diacritization and transliteration in a realistic setting characterized by ambiguity and spelling variation. We benchmarked GPT-4o on this task, providing insights into its capabilities and limitations. While the model performs reasonably well, especially on frequent names, it struggles with rarer entries and variant mappings.

Looking ahead, we plan to expand the dataset with more diverse names, integrate it into a morphological analyzer, and explore fine-tuned models for diacritizing proper nouns in broader contexts.
We also plan to fine-tune dedicated models for this task and develop more robust approaches to name ambiguity, especially with multiple valid diacritizations. We hope this resource advances Arabic NLP and name normalization in multilingual settings like Wikipedia.

\newpage
\newpage
\section*{Limitations}
A primary limitation of this work lies in the inherent subjectivity of diacritization, particularly for proper nouns where multiple correct variants may exist depending on regional, historical, or phonetic conventions. Despite rigorous annotation guidelines and quality checks, variability is an inevitable aspect of any human-annotated linguistic resource. 
Our current benchmark relies solely on GPT-4o, and we acknowledge the importance of evaluating performance across a broader range of large language models. While initial results are promising, the overall performance remains limited and, in our assessment, not yet suitable for reliable downstream use.

\section*{Ethics Statement}

All data used in this project were sourced from publicly available Arabic Wikipedia entries and their corresponding English titles, in accordance with Wikimedia’s terms of use. The annotation process was conducted transparently and ethically, with fair compensation provided to the annotators. We make both the corpus and the annotation guidelines publicly accessible under an open license, supporting reproducibility and community collaboration. Our goal is to contribute a valuable resource for Arabic language processing and to aid the broader Wikimedia effort by enhancing the quality of Arabic Wikipedia entries.
Finally, we acknowledge that all NLP tools and resources can be used with malicious intent; this is not our intention, and we categorically discourage it.

\section*{Benefits}

This work directly supports the Wikimedia community by enhancing the quality and accessibility of Arabic Wikipedia content. By providing more accurate diacritization for  proper nouns from all over the world on Arabic Wikipedia, we aim to improve readability, pronunciation, and downstream tasks such as named entity recognition and machine translation. The dataset, code, and annotation guidelines are all released under the Creative Commons Attribution-ShareAlike (CC BY-SA) license to ensure community reuse and adaptation. Filtering was applied to select single-word proper nouns related to people, locations, and organizations, drawn from Arabic Wikipedia entries that have clear English counterparts, thereby supporting multilingual alignment and cross-lingual research.

\section*{Risks}

Our project poses no known risks to Wikimedia editors or contributors. We do not name, identify, or reference any individual editor (by username or otherwise), nor do we expose any metadata that could be used to infer editor identities. The work focuses solely on content-level linguistic annotation and transformation. There are no known ways in which this research could be used to derive sensitive or personal information about contributors, and we strongly discourage any attempts to repurpose the resource for such purposes.

\section*{Acknowledgments}

We would like to express our sincere gratitude to Hamdy Mubarak for his valuable insights and his generous willingness to answer our questions throughout the course of this work. We thank Djellel Difallah for advice on initial data collection. We would like to also thank Bashar Alhafni and Mostafa Saeed for their insightful discussions and helpful conversations.

\bibliography{camel-bib-v3, custom, anthology}

\clearpage  
\appendix
\onecolumn
\section{GPT-4o Prompts}
\label{app:prompts}

In the system role, we provide the task description, and optionally, the few-shot demonstrations, when they are used.
For the user role, we always provide the single instance to be diacritized.
Table~\ref{tab:prompts} lists all of the prompts used for the different settings.
Table~\ref{tab:few-shot} shows a sample of the few-shot examples.
These are formatted as a markdown table in the prompts.

\pdfoutput=1

\begin{table*}[h!]
\footnotesize
\setlength{\tabcolsep}{2pt}
\centering
\begin{tabular}{lp{14cm}}

\toprule

\textbf{Shots} & \multicolumn{1}{c}{\textbf{Prompt}} \\

\toprule

& \multicolumn{1}{c}{\textbf{Arabic Word+Gloss Input}} \\
\midrule

Zero &

You are an expert in Arabic.

You are given the undiacritized proper noun in Arabic and its English gloss. 
Your task is to generate the corresponding diacritized proper noun lemma in Arabic.
Arabic lemmas are dictionary entries that have no attached definite article (\<ال>).
Diacritization is adding the correct diacritic markings to undiacritized words.

Remove the Arabic definite article (\<ال>) when present.
Do not add, remove, or substitute any other letters in the input.
Determine the most accurate diacritization that matches the English gloss pronunciation.

The user will provide a Markdown table with 1 rows.
Each row contains an undiacritized proper noun in Arabic in the “Input” column and its English gloss in the “Gloss” column.

Return exactly 1 diacritized lemmas, one per line.
Do not include extra text, explanations, or formatting. \\

\midrule

Few/One &

You are an expert in Arabic.

You are given the undiacritized proper noun in Arabic and its English gloss. 
Your task is to generate the corresponding diacritized proper noun lemma in Arabic.
Arabic lemmas are dictionary entries that have no attached definite article (\<ال>).
Diacritization is adding the correct diacritic markings to undiacritized words.

Remove the Arabic definite article (\<ال>) when present.
Do not add, remove, or substitute any other letters in the input.
Determine the most accurate diacritization that matches the English gloss pronunciation.

The user will provide a Markdown table with 1 rows.
Each row contains an undiacritized proper noun in Arabic in the “Input” column and its English gloss in the “Gloss” column.

Return exactly 1 diacritized lemmas, one per line.
Do not include extra text, explanations, or formatting.

Here are some examples of triplets of an undiacritized proper noun in Arabic (“Input”), its respective English gloss (“Gloss”), and its diacritized lemma (“Output”) for reference

\texttt{<Few-Shots-table>} \\

\bottomrule

& \multicolumn{1}{c}{\textbf{Arabic Word Only Input}} \\

\toprule

Zero & 

You are an expert in Arabic.

You are given the undiacritized proper noun in Arabic. 
Your task is to generate the corresponding diacritized proper noun lemma in Arabic.
Arabic lemmas are dictionary entries that have no attached definite article (\<ال>).
Diacritization is adding the correct diacritic markings to undiacritized words.

Remove the Arabic definite article (\<ال>) when present.
Do not add, remove, or substitute any other letters in the input.

The user will provide a Markdown table with 1 rows.
Each row contains an undiacritized proper noun in Arabic in the “Input” column.

Return exactly 1 diacritized lemmas, one per line.
Do not include extra text, explanations, or formatting. \\

\midrule

Few/One &

You are an expert in Arabic.

You are given the undiacritized proper noun in Arabic. 
Your task is to generate the corresponding diacritized proper noun lemma in Arabic.
Arabic lemmas are dictionary entries that have no attached definite article (\<ال>).
Diacritization is adding the correct diacritic markings to undiacritized words.

Remove the Arabic definite article (\<ال>) when present.
Do not add, remove, or substitute any other letters in the input.

The user will provide a Markdown table with 1 rows.
Each row contains an undiacritized proper noun in Arabic in the “Input” column.

Return exactly 1 diacritized lemmas, one per line.
Do not include extra text, explanations, or formatting.

Here are some examples of pairs of an undiacritized proper noun in Arabic (“Input”), and its diacritized lemma (“Output”) for reference

\texttt{<Few-Shots-table>} \\

\bottomrule

\end{tabular}
\caption{System prompts used in the experiments. \texttt{<Few-Shots-table>} is a placeholder for few-shot examples. In either setting, the user prompts consist solely of a single instance to be diacritized.}
\label{tab:prompts}
\end{table*}

\begin{table*}[ht]
\tabcolsep15pt
\centering
\begin{tabular}{rllrl}
\toprule
\multicolumn{2}{c}{\textbf{Arabic Word}}& \textbf{Gloss} &  \multicolumn{2}{c}{\textbf{Diacritized Reference}} \\
\midrule
\<ايدكس> & \textit{Aydks} & IDEX & \<إِيدِكس> & \textit{{\AHAMZADN}iydiks} \\
\<الغارديان> & \textit{AlgArdyAn} & Guardian & \<غَارْدِيَان> & \textit{gaAr.diyaAn} \\
\<رودريغيز> & \textit{rwdrygyz} & Rodriguez & \<رُودْرِيغِيز> & \textit{ruwd.riygiyz} \\
\<اوروغواي> & \textit{AwrwgwAy} & Uruguay & \<أُورُوغْوَاي> & \textit{{\AHAMZAUP}uwruwg.waAy} \\
\<بوتيه> & \textit{bwtyh} & Boutier & \<بُوتِيِه> & \textit{buwtiyih} \\
\<وايزمن> & \textit{wAyzmn} & Weizman & \<وَايزْمَن> & \textit{waAyz.man} \\
\bottomrule
\end{tabular}
\caption{A sample of few-shot examples used for prompting GPT-4o}
\label{tab:few-shot}
\end{table*}
\newpage 
\section{Supplementary Interplay of Frequency, Similarity, and Accuracy}
\label{app:freeman_frequency_analysis}

\begin{table}[ht]
\centering
\begin{tabular}{lrrrrrr}
\toprule
\textbf{Freeman Bin} & \textbf{Instances} & \textbf{Instance \%} & \textbf{Frequency} & \textbf{Matches} & \textbf{Accuracy} & \textbf{Distance} \\
\midrule
10\%   & 6     & 0.2\% & 2,280,059 & 5   & 83.3\% & 0.17 \\
20\%   & 7     & 0.2\% & 454,346   & 6   & 85.7\% & 0.29 \\
30\%   & 23    & 0.7\% & 303,728   & 20  & 87.0\% & 0.22 \\
40\%   & 27    & 0.8\% & 690,729   & 23  & 85.2\% & 0.26 \\
50\%   & 26    & 0.8\% & 64,814    & 23  & 88.5\% & 0.12 \\
60\%   & 71    & 2.1\% & 30,274    & 45  & 63.4\% & 0.69 \\
70\%   & 164   & 4.9\% & 57,361    & 124 & 75.6\% & 0.37 \\
80\%   & 271   & 8.1\% & 22,803    & 185 & 68.3\% & 0.46 \\
90\%   & 587   & 17.5\%& 22,909    & 404 & 68.8\% & 0.52 \\
100\%  & 2,180 & 64.8\%& 60,343    & 1,619 & 74.3\% & 0.38 \\
\midrule
10--90\%      & 1,182 & 35.2\% & {63,761}  & 835   & 70.6\% & 0.48 \\
\midrule
10--50\%      & 89    & 2.6\%  & {496,420} & 77    & 86.5\% & 0.20 \\
60--100\%     & 3,273 & 97.4\% & {49,719}  & 2,377 & 72.6\% & 0.42 \\
\midrule
All  & 3,362 & 100.0\% & 61,544    & 2,454 & 73.0\% & 0.41 \\
\bottomrule
\end{tabular}
\caption{Accuracy, average frequency, and edit distance across Freeman similarity score bins.}
\label{Freeman-Accuracy}
\end{table}

\begin{table}[h!]
\centering

\begin{tabular}{lccccc}
\toprule
\textbf{Frequency Range} & \textbf{Instances} & \textbf{Average Freq.} & \textbf{Matches} & \textbf{Accuracy} & \textbf{Avg. Freeman} \\
\midrule
Q1 (lowest 25\%)   & 787  & 2       & 510  & 64.8\% & 91.1\% \\
Q2 (25--50\%)      & 893  & 25      & 627  & 70.2\% & 90.4\% \\
Q3 (50--75\%)      & 840  & 567     & 646  & 76.9\% & 91.2\% \\
Q4 (highest 25\%)  & 842  & 245{,}145 & 671  & 79.7\% & 89.7\% \\
\midrule
All                & 3{,}362 & 61{,}544 & 2{,}454 & 72.99\% & 90.6\% \\
\bottomrule
\end{tabular}
\caption{Accuracy, Average Frequency, and average Freeman similarity scores across word frequency quartiles.}
\label{tab:frequency-accuracy}
\end{table}

\end{document}